\documentclass[10pt,twocolumn,letterpaper]{article}

\usepackage{iccv}
\usepackage{times}
\usepackage{epsfig}
\usepackage{graphicx}
\usepackage{amsmath}
\usepackage{amssymb}
\usepackage{kotex}

\usepackage{threeparttable}
\usepackage{commath}
\usepackage{bbm}
\usepackage{booktabs}
\usepackage{verbatim}
\usepackage{multirow}
\usepackage{bbm}
\usepackage{mathtools}
\usepackage{caption}
\usepackage{subcaption}
\usepackage[T1]{fontenc}
\usepackage[utf8]{inputenc}
\usepackage[table,xcdraw]{xcolor}

\newcommand*{\affmark}[1][*]{\textsuperscript{#1}}
\definecolor{MyGreen}{rgb}{0,0.6,0.3}
\definecolor{blackpink}{rgb}{0.6,0,0.6}
\definecolor{blood_red}{RGB}{200, 0, 0}

\newcommand\blfootnote[1]{%
  \begingroup
  \renewcommand\thefootnote{}\footnote{#1}%
  \addtocounter{footnote}{-1}%
  \endgroup
}

\newcommand{\specialcell}[2][c]{%
  \begin{tabular}[#1]{@{}c@{}}#2\end{tabular}}

\usepackage[pagebackref=true,breaklinks=true,letterpaper=true,colorlinks,bookmarks=false]{hyperref}

\iccvfinalcopy 

\def\iccvPaperID{7421} 

\ificcvfinal\fi

\begin{document}

\title{BiaSwap: Removing Dataset Bias with Bias-Tailored Swapping Augmentation}

\author{Eungyeup Kim\affmark[*]\\
KAIST\\
{\tt\small eykim94@kaist.ac.kr}
\and
Jihyeon Lee\affmark[*]\\
KAIST\\
{\tt\small jihyeonlee@kaist.ac.kr}
\and
Jaegul Choo\\
KAIST\\
{\tt\small jchoo@kaist.ac.kr}
}

\maketitle
\ificcvfinal\thispagestyle{empty}\fi

\begin{abstract}
Deep neural networks often make decisions based on the spurious correlations inherent in the dataset, failing to generalize in an unbiased data distribution.
Although previous approaches pre-define the type of dataset bias to prevent the network from learning it, recognizing the bias type in the real dataset is often prohibitive.
This paper proposes a novel bias-tailored augmentation-based approach, BiaSwap, for learning debiased representation without requiring supervision on the bias type.
Assuming that the bias corresponds to the easy-to-learn attributes, we sort the training images based on how much a biased classifier can exploits them as shortcut and divide them into bias-guiding and bias-contrary samples in an unsupervised manner.
Afterwards, we integrate the style-transferring module of the image translation model with the class activation maps of such biased classifier, which enables to primarily transfer the bias attributes learned by the classifier. 
Therefore, given the pair of bias-guiding and bias-contrary, BiaSwap generates the bias-swapped image which contains the bias attributes from the bias-contrary images, while preserving bias-irrelevant ones in the bias-guiding images.
Given such augmented images, BiaSwap demonstrates the superiority in debiasing against the existing baselines over both synthetic and real-world datasets.
Even without careful supervision on the bias, BiaSwap achieves a remarkable performance on both unbiased and bias-guiding samples, implying the improved generalization capability of the model.
\end{abstract}
\vspace{-0.3cm}
\blfootnote{* indicates equal contribution\vspace{-0.8cm}}

\section{Introduction}
\label{sec:introduction}
Recent deep neural networks have shown remarkable performances in computer vision tasks including classification and object detection.
However, these models often achieve their goals by erroneously relying on the peripheral features that have spurious correlations with their labels, so-called \textit{dataset bias}~\cite{datasetbias2011torralba}.
For instance, imagine a classifier for recognizing a \textit{camel}, when most of the camels in the training images appear in the desert.
This unintended correlation causes the classifier to overly rely on the attributes of the desert, failing to recognize the camel standing on the road.
In other words, the classifier trained on the biased dataset often shows drastic failures for the images without such bias, which raises a question on its generalization capability in unbiased image classification.

Existing methods attempt to address this issue by using an explicit definition of the bias type in their debiasing strategies.
Some approaches~\cite{wang2018hex, bahng2019rebias, geirhos2018imagenettrained} assume the texture bias in the image classification task and propose a hand-crafted module for such bias type.
Similarly, textual modality, \textit{i.e.,} question and answer, are pre-defined as the bias~\cite{rubi, clark-etal-2019-dont} in visual question answering task and resolved by leveraging the question-only network.

However, assuming an already-known bias is quite unrealistic in that bias attributes can vary according to the composition of the training dataset.
Moreover, unlike synthetic datasets where bias attributes are manually designated, \textit{e.g.,} color is set as bias in Colored MNIST as shown in Figure~\ref{fig:example}-(a), it is highly demanding to acquire the prior knowledge on the bias that inherently exists in the real-world dataset.
Therefore, an unsupervised debiasing with no definition in advance is an appropriate approach for learning generalized representations over various datasets.
Moreover, maintaining the classification ability for biased samples as well as unbiased samples needs to be considered crucial for a desirable representation, which has been overlooked by the previous studies~\cite{clark-etal-2019-dont, nam2020learning}.

This paper mainly focuses on 1) removing the dataset bias without explicit supervision by leveraging the bias-tailored swapping augmentation, and 2) achieving superior performance on bias-contrary as well as bias-guiding samples against other baselines. 
We propose an image translation-based augmentation framework, \textit{BiaSwap}, which transfers the attributes appearing on the regions of the image where the classifier often exploits as a shortcut for prediction.
To this end, we first exploit the reasonable observation that the biased classifier often learns to exploit \textit{easy-to-learn} attributes in the early learning phase, proposed in Nam \etal~\cite{nam2020learning}
This enables us to obtain the class activation map (CAM)~\cite{zhou2015cam} which indicates the bias-relevant regions for each image without requiring an explicit definition of the bias type in advance.
By integrating the CAM into the image translation framework, we augment the images with their bias attributes being translated by those of another exemplar image.
At the same time, we present a simple and intuitive criterion based on the same assumption (\textit{i.e.,} bias is \textit{easy-to-learn}) for discriminating between the bias-guiding and the bias-contrary samples among the training set.
Therefore, given the pairs, BiaSwap mainly translates the bias-guiding image into the bias-contrary one by transferring the specific attributes corresponding to the bias.
These augmented images, termed as \textit{bias-swapped}, make the proportion of bias-guiding images to be less dominant in the training dataset, removing the dataset bias in the end.
We provide extensive experiments representing that BiaSwap achieves the state-of-the-art debiasing results against the baselines across various datasets from synthetic (\textit{i.e.,} Colored MNIST, Corrupted CIFAR10) to real-world (\textit{i.e.,} BAR, bFFHQ) datasets, even without explicit supervision on the bias type.

\section{Preliminaries}
In this section, we first provide the formulation of the dataset bias (Section~\ref{sec: bias definition}).
Afterward, we categorize the various existing debiasing approaches in terms of prior assumption on bias type (Section~\ref{sec: related work}).

\subsection{Definition of unwanted correlation in dataset}
\label{sec: bias definition}
Consider a training dataset $\mathcal{D}$ where each image $x\in\mathcal{X}$ has its corresponding class label $y\in\mathcal{Y}$.
Each $x$ can be explained by its various visual attributes, such as shape and color, and some of them are exploited by a classifier in the image classification task.
Among these attributes, let $z_g$ the one that is essential for predicting a target label $y$, meaning that every image for class $y$ must contain $z_g$.
Therefore, a classifier becomes generalized in the unbiased distribution when learning this attribute as a cue.
In contrast, let $z_b$ denote the attribute which is less essential, but have a strong correlation with target label $y$.
In addition, $z_b$ often acts as the bias attribute when it is easier for the classifier to learn compared to $z_g$.
Eventually, the model becomes biased by overly exploiting the $z_b$ instead of $z_g$ when trained in the biased dataset, failing to predict the samples which do not contain the $z_b$.
For example, in the Colored MNIST, most of the images in each class are highly correlated with the specific color, as illustrated in Figure~\ref{fig:example}-(a).
On the other hand, an unbiased test set contains the samples whose colors are uniform at random, having no correlation with their target label.
In this case, attributes $z_g$ corresponds to the digit, while $z_b$ indicates the color in each image.
Throughout the paper, we term $z_b$ \textit{bias-guiding} attribute and the image containing the $z_b$ as bias-guiding image. 
While most of the samples with the same class in the training distribution share the $z_b$, there might be a small portion of samples that have attributes that are conflicting to $z_b$, which we term $z_{-b}$.
For example, in the Colored MNIST, while most of the samples in class $0$ contain \textit{red}, a few samples contain \text{non-red} color, such as blue or green.
As this $z_{-b}$ attribute is contradictory against $z_{b}$, the biased network cannot rely on it anymore.
We term $z_{-b}$ \textit{bias-contrary} attribute, and the image with $z_{-b}$ as bias-contrary image.

Since the bias-guiding samples with $z_b$ are dominant in the training dataset, it leads the classifier to rely on $z_b$ rather than the essential attribute $z_g$.
Therefore, removing the dataset bias by increasing the proportion of the bias-contrary samples with $z_{-b}$ can encourage the model to learn $z_g$ by preventing it from relying solely on the $z_b$ for classification.
Our proposed image translation-based augmentation approach generates the images with their visual aspects of $z_b$ being transferred into $z_{-b}$, while maintaining the essential features $z_g$.
We term this augmented sample as a \textit{bias-swapped} image.
This, as a result, leads our classifier to achieve consistent performance in unbiased dataset distribution, where most of the samples are bias-contrary.

\subsection{Existing debiasing approaches}
\label{sec: related work}
\noindent{\textbf{Remove bias with prior knowledge}}
Several approaches with an explicit label on the bias type have been proposed~\cite{Kim_2019_CVPR,li2019repair,Agarwal2020TowardsCV, sagawa2019distributionally, goel2021model}.
Li and Vasconcelos~\cite{li2019repair} and Kim \etal~\cite{Kim_2019_CVPR} set the particular RGB values to be a bias cue in the Colored MNIST dataset, where a specific color is correlated with each digit.
Agarwal \etal~\cite{Agarwal2020TowardsCV} propose to synthesize the data with a generative algorithm by involving manually curated heuristics which selects the objects to remove. 
Besides, Sagawa \etal~\cite{sagawa2019distributionally} and Goel \etal~\cite{goel2021model} utilize the clustering of the bias subgroups which require expensive supervision on the bias type.
 
Other approaches pre-define the bias type in advance and build a bias-tailored module for addressing the certain bias type~\cite{bahng2019rebias,wang2018hex,geirhos2018imagenettrained,rubi,clark-etal-2019-dont,Shetty2019NotUT,shahcylce, Ray2019SunnyAD}.
Wang \etal~\cite{wang2018hex} assume the texture bias in the image classification task, and propose a projection method in the latent space to learn the independent features from the texture-biased ones.
Geirhos \etal~\cite{geirhos2018imagenettrained} propose a style transfer-based augmentation method with adaptive instance normalization~\cite{huang2017adain}, which enhances the robustness against the texture bias. 
Bahng \etal~\cite{bahng2019rebias} introduce the model with limited capacity for capturing the texture bias in image classification or static bias in video action recognition, respectively, and propose the learning of the statistically independent representation against it. 

However, these approaches have limitations in that assuming the certain type of bias does not guarantee the generalized debiasing in the dataset with other types of bias. 
As the bias-guiding attributes $z_b$ is determined by the characteristics of dataset, such as the composition of images and the attribute complexity, learning debiased representation without prior assumption on the certain bias type is essential.

\noindent{\textbf{Remove bias without explicit supervision}}
Still, learning debiased representation in an unsupervised manner is an ideal but demanding problem.
Darlow \etal~\cite{darlow2020latent} utilize the adversarial perturbation in the latent space for synthesizing the images against the bias that the classifier learns.
Nam \etal~\cite{nam2020learning} observe the general aspect of bias as easy-to-learn in the early training phase and adopt the generalized cross-entropy loss~\cite{zhang2018generalized} to train a biased network.
The samples in which such biased network fails to classify are then emphasized through weighted cross-entropy loss in the training of the debiased network.

A truly debiased classifier learns the generalized attribute $z_g$, which should correctly classify the samples in the unbiased as well as the biased dataset.
However, existing baselines~\cite{clark-etal-2019-dont, nam2020learning} often suffer from the significant performance degradation in the biased dataset (\textit{i.e.}, bias-guiding samples).
This implies that they implicitly learn to \textit{avoid} the bias-guiding attributes, not fully learning the $z_g$.
Learning debiased representation without bias supervision remains challenging, and is thus fairly under-explored.
Our proposed bias-tailored augmentation effectively removes the dataset bias, achieving the generalized debiasing capability in both biased and unbiased test set.

\section{Proposed Approach}
\label{sec: proposed approach}
This section provides a detailed description of distinguishing a bias-guiding and bias-contrary samples (Section~\ref{sec:3.1}), training a bias-tailored swapping autoencoder (Section~\ref{sec:bias-tailored swapping autoencoder}), and training a classifier with debiased representation (Section~\ref{sec:3.3}).

\subsection{Separation of bias-contrary samples}
\label{sec:3.1}
We propose a simple, yet effective method that divides the training samples into bias-guiding and bias-contrary groups.
Our method assigns the bias label for the images adaptively according to the training dataset, without the explicit supervision on the bias.
As mentioned in Section~\ref{sec: bias definition}, bias-guiding samples have the unwanted correlations that are \textit{easy-to-learn}~\cite{nam2020learning, geirhos2020shortcut}, while bias-contrary samples are \textit{hard-to-learn}.
Therefore, a biased classifier becomes \textit{certain} and \textit{correct} for the bias-guiding samples.
In contrast, as the bias-contrary samples do not include the attributes the classifier mainly relies on, the classifier can be either 1) \textit{certain} and \textit{incorrect} or 2) \textit{uncertain} for the bias-contrary samples.
Based on these characteristics, we introduce a \textit{pseudo-bias label} which determines whether each image belong to bias-guiding or bias-contrary by observing both the classification correctness and the confidence of the model for the image.

\begin{figure*}
\begin{center}
\includegraphics[width=\linewidth]{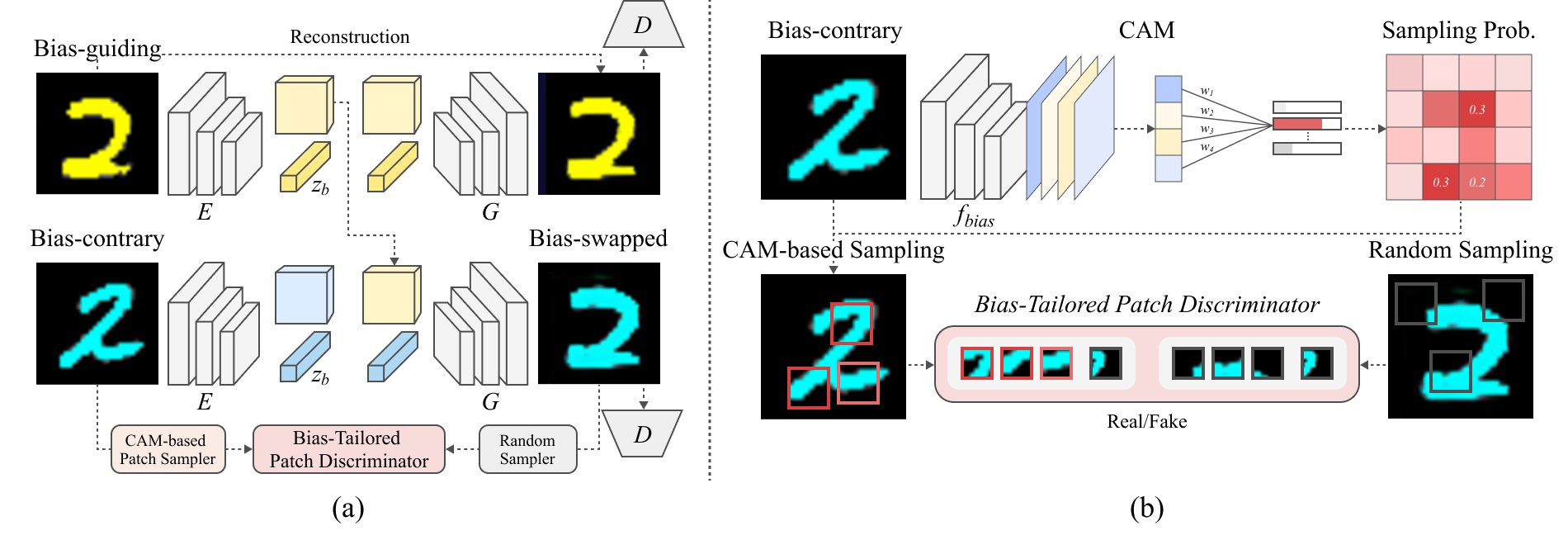}
\end{center}
\vspace{-0.5cm}
\caption{Illustration of the proposed method, BiaSwap. The figure (a) shows the overall pipeline of the swapping augmentation framework and the figure (b) describes the patch samplers and bias-tailored patch discriminator in detail. 
We generate the bias-swapped images from this framework to augment the training dataset for learning debiased representation.}
\label{fig:main_fig}
\vspace{-0.5cm}
\end{figure*}

To distinguish the binary category (\textit{i.e.,} bias-contrary or bias-guiding) more certainly, we first train the biased classifier $f_\text{bias}$ by exploiting the generalized cross-entropy (GCE) loss~\cite{zhang2018generalized} in a similar manner to Nam \etal~\cite{nam2020learning}
GCE loss is originally proposed as a noise-robust alternative to the categorical cross-entropy (CE) loss. 
In our setting, it amplifies the biased representation since its gradient is written as $\frac{\partial GCE(p,y)}{\partial \theta} = \frac{{p_y^q} \partial CE(p,y)}{\partial \theta}$ where $p_y$ is probability corresponding to the target label $y$, $q \in (0,1]$ is a hyperparameter, and $\theta$ denotes the network parameters. 
GCE loss puts greater importance on the samples which are easy to learn compared to the CE loss. 
As these samples are bias-guiding in our training dataset, our classifier becomes biased.

Assume the biased classifier $f_\text{bias}$ outputs the resulting logits $\textbf{z}=(z_1,...,z_K) \in \mathbb{R}^K$ after the last linear layer, where $K$ denotes the number of the target class.
We first define the bias score of each sample $x$ 
by obtaining the absolute difference between the correctness and the max values of probability described as
\begin{equation}
    \begin{split}
    \text{score}(x)&=
    \bigg\lvert{
    \mathbbm{1}_{\text{argmax}_{k} f_\text{bias}(x)=y}
    - \texttt{max}\Big(\frac{e^{f_\text{bias}(x)}}{\sum_{j=1}^K{e^{f_\text{bias}(x)_j}}}\Big) 
    \bigg\rvert},
    \end{split}
\label{eq:bias score}
\end{equation}
where $\mathbbm{1}$ is an indicator function which outputs one when satisfying the given condition and zero in vice versa, \texttt{max} returns the maximum probability value after the softmax operation, and $f_\text{bias}(x)_j$ as the $j$-th logit values of the biased classifier.
For the bias-guiding image which contains $z_b$, the model correctly predicts the target label $y$ (\textit{i.e.,} first term in Eq.~\ref{eq:bias score} becomes 1) with high confidence (\textit{i.e.,} second term in Eq.~\ref{eq:bias score} becomes high), resulting the calculated score to be close to $0$.
Contrariwise, the score becomes close to $1$ when the model makes the wrong prediction (\textit{i.e.,} first term in Eq.~\ref{eq:bias score} becomes $0$) with high confidence (\textit{i.e.,} second term in Eq.~\ref{eq:bias score} becomes high), which might be mostly observed for the bias-contrary samples.
In addition, for the occasional cases where the classifier correctly predicts the bias-contrary images with low confidence, the score would be placed between $0$ and $1$.
Given such score, we determine the pseudo bias label $\tilde{y}_{\text{bias}}(x)$ for each data which is
\begin{equation}
    \begin{split}
    \tilde{y}_{\text{bias}}(x)&=\begin{cases}
      1 & \text{if}\ \ \text{score}(x) > \frac{1}{N}\sum^N_{i=0} \text{score}(x_i) \\
      0 & \text{otherwise}
    \end{cases}
    ,
    \end{split}
\label{eq:pseudo bias label}
\end{equation} where $y$ denotes a ground-truth target label and $N$ as the total number of training images.
We consider the image assigned with $\tilde{y}_\text{bias}=0$ as bias-guiding and that with $\tilde{y}_\text{bias}=1$ as bias-contrary.
We adopt the arithmetic mean of the scores over the entire samples as the threshold for determining whether each sample is bias-guiding or bias-contrary. 
We empirically find that such mean values of the scores can act as a simple and effective threshold for discriminating the bias-guiding images and bias-contrary images.
In Section~\ref{sec:Quantitative}, we validate this simple criterion reasonably performs over the various datasets utilized in the paper. 

\subsection{Bias-tailored swapping autoencoder}
\label{sec:bias-tailored swapping autoencoder}
Given the pair of bias-guiding and bias-contrary images using the $\tilde{y}_\text{bias}$ as shown in Figure~\ref{fig:main_fig}-(a), we leverage the state-of-the-art image-to-image translation method called swapping autoencoder (SwapAE)~\cite{park2020swapping} as our backbone network for translation.
To enable the translation of the bias-aware attributes in the bias-aligned samples to be bias-contrary, we propose a novel variant of patch cooccurrence discriminator, which mainly focuses on bias attributes based on the class activation map (CAM)~\cite{zhou2015cam} of a biased classifier. 

\noindent{\textbf{Swapping autoencoder}}
\label{sec:swapping autoencoder}
SwapAE~\cite{park2020swapping} consists of the encoder $E$ which maps the image into the latent features $z$, and the generator $G$ which reconstructs the images $x$ from $z$.
Specifically, $E$ encodes the image into its content features $z_c$ and style features $z_s$, and $G$ takes them to synthesize the image which can be explained by them.
SwapAE first utilizes the reconstruction loss and adversarial loss~\cite{goodfellow2014gans} for generating both realistic reconstruction of an input image $x$.
Both losses are written as
\begin{equation}
    \begin{split}
        \mathcal{L}_{\text{recon}}(E,G)&=\mathbb{E}_{x\sim \mathcal{X}}\left[\big\|x - G(E(x))\big\|^2_2\right],\\
        \mathcal{L}_{\text{GAN,recon}}(E,G,D)&=\mathbb{E}_{x\sim \mathcal{X}}\left[-\text{log}D(G(E(x)))\right],
    \end{split}
\end{equation}
where $\mathcal{X}$ denotes a training dataset distribution and $D$ a discriminator which classifies whether the image is real or fake.
In addition, SwapAE learns to translate the style of an image, denoted as $x^1$, into that of another one, $x^2$, generating the translated image.
This can be done by constructing the swapped pair of latent features from these images and decoding them into the image.
In other words, each pair of ($z_c^1$, $z_s^1$) and ($z_c^2$, $z_s^2$) are encoded from $x^1$ and $x^2$, respectively, and the swapped pair, \textit{i.e.,} ($z_c^1$, $z_s^2$), are decoded to generate the translated image, which contains the style of $x^2$ while maintaining the content of $x^1$.
To ensure the translated images with swapped attributes to contain the same style with $x^2$, a patch cooccurrence discriminator $D_{\text{patch}}$ is proposed.
Such discriminator enforces the styles in the randomly sampled patches from the generated images to be identical to the ones in $x^2$.
Therefore, the objective function can be written as
\begin{equation}
\label{eq:patch_cooccurence}
    \begin{split}
        &\mathcal{L}_{\text{CooccurGAN}}(E,G,D_{\text{patch}})=\\
        &\mathbb{E}_{x^1,x^2\sim \mathcal{X}} \left[-\text{log}(D_{\text{patch}}(\small\texttt{crop}_\texttt{u}(G(z^1_c, z^2_s)), \small\texttt{crops}_\texttt{u}(x^2)))\right],
    \end{split}
\end{equation}
where $\texttt{crop}_\texttt{u}$ and $\texttt{crops}_\texttt{u}$ denote the operation of cropping \textit{uniformly at random} in an image for a single patch and multiple patches, respectively.
To make the generated images $G(z^1_c, z^2_s)$ realistic, the adversarial loss is added as
\begin{equation}
    \begin{split}
        &\mathcal{L}_{\text{GAN,swap}}(E,G,D)=\\
        &\mathbb{E}_{x^1,x^2\sim \mathcal{X},x^1\neq x^2} \left[-\text{log}(D(G(z^1_c, z^2_s))\right].
    \end{split}
\end{equation}

\noindent{\textbf{CAM-based patch sampling}}
\label{sec:cam-based patch sampling}
As $D_{\text{patch}}$ randomly samples the patches from the entire spatial resolution, the style extracted from the patches does not reflect certain attributes.
Instead, as we aim to transfer the styles corresponding to the attributes the classifier easily learns as shortcuts, sampling the patches related to these attributes are required.
Therefore, we leverage the biased classifier $f_\text{bias}$ and integrate its CAM~\cite{zhou2015cam}, which identifies the discriminative regions used by such a classifier, into the patch sampling method in $D_\text{patch}$.
Specifically, given an image, our classifier produces an activation map $f_\text{bias,k}(x,y)$, where k denotes a channel index and $(x,y)$ the coordinate for the spatial location.
Then, following Zhou \etal~\cite{zhou2015cam}, we calculate a logit for each class $c$ as $\sum_k{w^c_k}F_k$, where $F_\text{bias,k}$ denotes the result of global average pooling for channel $k$ and $w^c_k$ indicates a weight which maps the $F_\text{bias,k}$ into each class probability. 
The logits for class $c$ can be written as 
\begin{equation}
    \begin{split}
        \sum_k{w^c_k}F_\text{bias,k} &= \sum_k{w^c_k}\sum_{x,y}f_\text{bias,k}(x,y)\\
        &=\sum_{x,y}\sum_{k}{w^c_k}f_\text{bias,k}(x,y).
    \end{split}
\end{equation} Therefore, the importance of the activation map at spatial location $(x,y)$ for classifying the class $c$ by the classifier $f_\text{bias}$ can be presented as
\begin{equation}
    I_c(x,y) = \sum_{k}{w^c_k}f_\text{bias,k}(x,y).
\end{equation}
As the classifier is biased, the large value of $I_c(x,y)$ demonstrates the location where the bias attributes are highly obtained by the classifier.
In this regard, we convert $I_c(x,y)$ into the sampling probability $P{(x,y)}$ for each spatial location of patch $(x,y)$ and utilize such probability in the discriminator $D_\text{patch}$ for style extraction, as shown in Figure~\ref{fig:main_fig}-(b).
In other words, instead of random \texttt{crop} operation in Eq.~\ref{eq:patch_cooccurence}, we utilize the cropping of local patches according to the probability described as
\begin{equation}
\label{eq:bias_sampling}
    \begin{split}
        P{(x,y)}=\frac{e^{I_c(x,y)}}{\sum_{x,y} e^{I_c(x,y)}}.
    \end{split}
\end{equation} 
This encourages the more frequent sampling of patches corresponding to the bias attribute in the images compared to others, enabling the translation of bias attributes from an image to another.
Therefore, the variant of the objective function of Eq.~\ref{eq:patch_cooccurence} via bias-tailored patch discriminator can be described as
\begin{equation}
\label{eq:bias-tailored_patch_cooccurence}
    \begin{split}
        &\mathcal{L}_{\text{CooccurGAN}}(E,G,D_{\text{bias-tailored patch}})=\\
        &\mathbb{E}_{x^1,x^2\sim \mathcal{X}} \left[-\text{log}(D_{\text{patch}}(\small\texttt{crop}_\texttt{b}(G(z^1_c, z^2_s)), \small\texttt{crops}_\texttt{b}(x^2)))\right],
    \end{split}
\end{equation}
where $\texttt{crop}_\texttt{b}$ and $\texttt{crops}_\texttt{b}$ denote the operation of cropping under the probability of Eq.~\ref{eq:bias_sampling} in an image for a single patch and multiple patches, respectively.
In consequence, BiaSwap generates the \textit{bias-swapped} image, which contains the bias-relevant attributes from the bias-contrary image while preserving the bias-irrelevant features from the bias-guiding image, as shown in Figure~\ref{fig:main_fig}-(a).

\subsection{Training classifier with augmented dataset}
\label{sec:3.3}
By adding the generated bias-swapped images $\mathcal{X}_\text{bias-swapped}$,
we can obtain our augmented training dataset $\mathcal{X}_\text{aug}=\mathcal{X}\cup\mathcal{X}_\text{bias-swapped}$.
These reasonable amounts of bias-swapped samples in $\mathcal{X}$ alleviates the dataset bias caused by the dominant number of bias-guiding images in the dataset, thus preventing the model from learning biased representation.
Finally, we train a classifier $f_\text{debias}$ with these datasets with the classification loss as
\begin{equation}
    \mathcal{L}_\text{class}=\mathbb{E}_{x\sim \mathcal{X_{\text{aug}}}}\left[-\sum_{c}y_c\text{log}f_\text{debias}(x)\right].
\end{equation}

\section{Experiments and Analysis}
In Section~\ref{sec:experimental setup}, we first introduce the experimental setup, including the details of the biased datasets and implementation details. 
Afterward, Section~\ref{sec:Quantitative} and Section~\ref{sec: qualitative} provide the quantitative and qualitative comparison between our method with existing baselines on synthetic and real-world datasets, respectively. 

\subsection{Experimental setup}
\label{sec:experimental setup}

\begin{figure}
\centering
\begin{subfigure}[b]{\linewidth}
    \includegraphics[width=\linewidth]{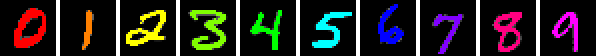}
    \vspace{-0.5cm}
    \caption{Colored MNIST}
\end{subfigure}
\begin{subfigure}[b]{\linewidth}
    \includegraphics[width=\linewidth]{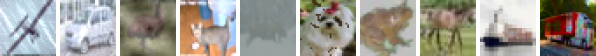}
    \vspace{-0.5cm}
    \caption{Corrupted CIFAR10}
\end{subfigure}
\begin{subfigure}[b]{\linewidth}
    \includegraphics[width=\linewidth]{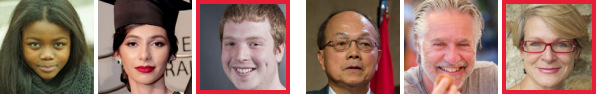}
    \vspace{-0.5cm}
    \caption{bFFHQ}
\end{subfigure}
\begin{subfigure}[b]{\linewidth}
    \includegraphics[width=\linewidth]{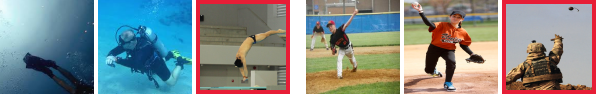}
    \vspace{-0.5cm}
    \caption{BAR}
\end{subfigure}
    \vspace{-0.5cm}
  \caption{Example images of each dataset we utilize in the paper. 
  Rows (a) and (b) represent the bias-guiding samples which have a strong correlation between bias attribute and the target class.
  For rows (c) and (d), we additionally visualize the bias-contrary images with red boxes, which do not contain such correlation.
  }
\label{fig:example}
\vspace{-0.5cm}
\end{figure}

We evaluate our method as well as the baselines across the synthetic dataset, \textit{i.e.,} Colored MNIST, and Corrupted CIFAR10~\cite{hendrycks2018benchmarking}, which are widely used in the previous literature.
We also utilize the real-world datasets including BAR~\cite{nam2020learning} and bFFHQ.

\begin{table*}[t!]
\begin{center}
\setlength\tabcolsep{4.5pt}
\begin{tabular}{c|c|ccc||c|ccc||c}
\toprule
\multirow{2}{*}{Dataset} & \multirow{2}{*}{\%} & \multicolumn{4}{c|}{Bias-guiding} & \multicolumn{4}{c}{Unbiased} \\ &&\cellcolor[HTML]{EFEFEF}Vanilla & \cellcolor[HTML]{FFF2B7}ReBias & \cellcolor[HTML]{DCEDFF}LfF & \cellcolor[HTML]{DCEDFF}BiaSwap & \cellcolor[HTML]{EFEFEF}Vanilla & \cellcolor[HTML]{FFF2B7}ReBias & \cellcolor[HTML]{DCEDFF}LfF & \cellcolor[HTML]{DCEDFF}BiaSwap \\ 
\midrule \midrule
\multirow{4}[4]{*}{Colored MNIST} &95.0 & 99.23 & \textbf{100.0} & 80.33 & \underline{97.95} & 79.54 & \textbf{96.28} & 84.72 & \underline{90.85} \\
\cmidrule{2-10}
&98.0 &99.80 & \textbf{100.0} & 69.14 & \underline{98.33} & 62.62 & \textbf{90.16} & 75.88 & \underline{85.29} \\
\cmidrule{2-10}
&99.0 & 99.74 & \textbf{100.0} & 62.33 & \underline{98.45} & 48.76 & \textbf{84.19} & 70.05 & \underline{83.74} \\
\cmidrule{2-10}
&99.5 & 99.44 & \textbf{99.9} & 72.85 & \underline{98.49} & 32.67 & 62.82 & 61.61 & \underline{\textbf{85.76}} \\
\toprule
\multirow{4}[4]{*}{Corrupted CIFAR10} &95.0 & \textbf{99.05} & 98.23 & 62.74 & \underline{95.53} & 35.68 & \textbf{45.49} & \underline{42.32} & 41.62 \\
\cmidrule{2-10}
&98.0 & \textbf{98.97} & 98.67 & 73.41 & \underline{94.82} & 29.68 & 31.52 & 35.23 & \underline{\textbf{35.25}} \\
\cmidrule{2-10}
&99.0 & 98.79 & \textbf{99.11} & 74.73 & \underline{96.98} & 24.51 & 25.04 & 29.27 & \underline{\textbf{32.54}} \\
\cmidrule{2-10}
&99.5 & 98.56 & \textbf{99.29} & 80.90 & \underline{96.82} & 23.12 & 20.49 & 27.10 & \underline{\textbf{29.11}} \\
\bottomrule
\end{tabular}
\end{center}
\vspace{-0.5cm}
\caption{Quantitative comparisons of bias-guiding and unbiased test accuracy on two synthetic datasets. Note that each method has a different supervision level.
The methods with yellow background assume the bias type in advance, while those with blue do not require such type. We denote the best score with bold and the best score among unsupervised methods with under-lined scores.}
\label{tab: sythetic}
\vspace{-0.5cm}
\end{table*}

\noindent\textbf{Datasets}
As shown in Fig.~\ref{fig:example}, Colored MNIST is an MNIST dataset~\cite{lecun-mnisthandwrittendigit-2010} which has a correlation with certain colors. To inject the color bias, we select 10 distinct colors and inject each color into the MNIST images with a certain digit label (\textit{e.g.,} red color for images of zero label). Bias-contrary samples have their colors sampled uniformly at random.
Corrupted CIFAR10 is the CIFAR10~\cite{Krizhevsky09} dataset with texture corruptions, as proposed in Hendrycks and Dietterich~\cite{hendrycks2018benchmarking}. Similar to Colored MNIST, each texture corruption has an injurious correlation with each object class. 
We newly construct the Gender-biased FFHQ dataset (bFFHQ) which has age as a target label and gender as a correlated bias, and the images are from the FFHQ dataset~\cite{karras2019style}. The images include the dominant number of young women (\textit{i.e.}, aged 10-29) and old men (\textit{i.e.}, aged 40-59) in the training data. 
The biased action recognition (BAR) dataset is categorized by six human-action classes that are correlated with the distinct places~\cite{nam2020learning}. 
Curated six typical action-place pairs are \textit{(Climbing, RockWall), (Diving,
Underwater), (Fishing, WaterSurface), (Racing, A PavedTrack), (Throwing, PlayingField)}, and \textit{(Vaulting, Sky)}.
For the experiments on synthetic datasets, we vary the ratio of bias-guiding samples which are 95.0\%, 98.0\%, 99.0\%, and 99.5\%.
For bFFHQ, we utilize 99.0\% of bias-guiding images.
For BAR, we utilize typical action-place paired images for training, and bias-contrary ones only belong to the evaluation set. 
Although BAR only contains the bias-guiding training samples, undoubtedly there exist relatively easier samples, \textit{i.e.,} more bias-guiding, than the others \textit{i.e.,} less bias-guiding, in our proposed framework. 

\noindent\textbf{Evaluation sets}
To measure the generalization capability of the debiasing method, we consider the two types of evaluation sets, the unbiased and bias-guiding sets.
The unbiased evaluation set is constructed in a way that the bias attributes are distributed uniformly at random among the data without any correlation with a certain target label, following the evaluation protocol of existing studies~\cite{bahng2019rebias, nam2020learning, darlow2020latent}.
This set mainly evaluates how the debiasing method correctly classifies the bias-contrary test samples which do not include the strong correlation.
Note that for the real-world datasets, we exclude the bias-guiding samples from the unbiased test set, and call the remaining ones as ``bias-contrary" test set, as did in LfF~\cite{nam2020learning}.
In contrast, the bias-guiding set is composed of the bias-guiding images from the same distribution of the biased training dataset.
Such an evaluation set enables us to evaluate how the debiasing method maintains the classification capability for the bias-aligned test images after learning the debiased representation.
We believe that the truely debiased classifier should correctly predict the target labels of images in an unbiased as well as a biased test sets.

\noindent{\textbf{Implementation details}}
For the biased classifier and de-biased classifier, we use MLP with three hidden layers for Colored MNIST, and ResNet-18~\cite{He2015resnet} for  Corrupted CIFAR10, bFFHQ, BAR datasets, respectively.
For the SwapAE, we follow the same network architecture as proposed in Park \etal~\cite{park2020swapping}
To measure the bias score, hyperparameter $q=0.7$ is used for the GCE loss, and thresholds are fairly chosen on 50 epochs.
We provide a detailed description of experimental details in Section~\ref{sec:implementation} in the supplementary material.

\noindent{\textbf{Comparison methods}}
We compare BiaSwap with existing methods, ReBias~\cite{bahng2019rebias} and LfF~\cite{nam2020learning}, which address the dataset bias problem in the image classification task. 
The vanilla classifier which is trained without any debiasing procedure is also included in the comparison.
In addition, we add Stylised ImageNet (SIN)~\cite{geirhos2018imagenettrained} as our baseline in validating the effectiveness of utilizing the realistic augmentation in debiasing.
For the fair comparison with baseline models, we re-implement LfF~\cite{nam2020learning}, ReBias~\cite{bahng2019rebias}, and Stylised ImageNet (SIN)~\cite{geirhos2018imagenettrained} with the dataset we evaluated. 

\subsection{Quantitative Evaluation}
\label{sec:Quantitative}

\begin{table}[t!]
\begin{center}
\scalebox{0.84}{
\begin{tabular}{c|c|c|c|c||c}
\toprule
\multicolumn{2}{c|}{Dataset} & \multicolumn{1}{c|}{Vanilla} & \multicolumn{1}{c|}{ReBias} & \multicolumn{1}{c||}{LfF} & \multicolumn{1}{c}{BiaSwap} \\
\midrule
\midrule
\multirow{2}[1]{*}{bFFHQ} & Bias-guiding & 98.60 & 98.09 & 59.85 & \underline{\textbf{99.13}} \\
\cmidrule{2-6}
 & Bias-contrary & 51.03 & 53.66 & 55.61 & \underline{\textbf{58.87}} \\
\toprule
\multirow{2}[1]{*}{BAR} & Bias-guiding & \textbf{95.00} & 87.78 & 91.67 & \underline{93.33} \\
\cmidrule{2-6}
 & Bias-contrary & 49.59 & 39.29 & 52.13 & \underline{\textbf{52.4}} \\
\bottomrule
\end{tabular}
}
\end{center}
\vspace*{-0.5cm}
\caption{Quantitative comparisons of bias-guiding and bias-contrary test accuracy on two real-world datasets.
We denote the best score with bold and the best score among unsupervised methods with under-lined scores.}
\label{tab: real world dataset}
\vspace*{-0.5cm}
\end{table}

\begin{table}[!b]
\vspace{-0.4cm}
\begin{center}
\scalebox{0.88}{
\begin{tabular}{c|cccccc}
\toprule
Dataset && \specialcell{Colored \\ MNIST} & \specialcell{Corrupted \\ CIFAR10} & bFFHQ \\
\midrule \midrule
\specialcell{Precision (\%)} && 97.54 & 60.70 & 65.52\\
\cmidrule{1-5}
\specialcell{Recall (\%)} && 92.12 & 87.28 & 70.62\\
\cmidrule{1-5}
F1 score (\%) && 94.74 & 66.13 & 67.70\\
\bottomrule
\end{tabular}
}
\vspace{-0.1cm}
\end{center}
\vspace{-0.3cm}
\caption{Quantitative evaluations on the $\tilde{y}_\text{bias}$ assignment via precision, recall, and F1 score metrics. We report the evaluation scores for 99.0\% of each dataset, except BAR where no bias label is accessible.}
\label{tab: confusing score}
\vspace{-0.3cm}
\end{table}

\noindent{\textbf{Synthetic datasets}}
We verify our method on Colored MNIST and Corrupted CIFAR10.
In Table~\ref{tab: sythetic}, we report the classification accuracy on the two datasets with varying ratios of bias, evaluated on both bias-guiding and unbiased test set for each dataset. 
The more bias becomes severe, the more vanilla model failure to generalize on unbiased data, where the shortcut does not exist. 
In contrast, our proposed method maintains the robust debiasing capability on the unbiased test set, regardless of the bias ratio.
In addition, the ReBias is observed to achieve the best score in Colored MNIST, which is mainly due to the network designed for capturing the texture bias.
Note that BiaSwap obtains comparable accuracies compared to ReBias, even BiaSwap does not require any assumption on the type of bias in advance.
Especially for the bias-guiding samples, our method achieves significant improvement, as success to generalize on the intended direction.
Compared to LfF, where no prior knowledge on the bias type is required like BiaSwap, BiaSwap outperforms LfF on both bias-guiding and unbiased test accuracies for most of the dataset setup.
As for the degraded bias-guiding accuracies of LfF, we assume that the oversampling of the limited number of bias-contrary images makes the network instead under-fitted to the downplayed bias-guiding images during training.

\noindent{\textbf{Real-world datasets}}
To demonstrate the efficacy of our method on a realistic scenario, we provide the quantitative comparisons validated on bFFHQ and BAR datasets that contain complex types of bias in real-world images.
Table~\ref{tab: real world dataset} demonstrates that BiaSwap achieves superior debiasing performance against the existing baselines on these real-world datasets.
ReBias reveals the significant drop of bias-contrary accuracy for those datasets where the texture no more causes the unwanted correlation.
Likewise, LfF shows the degraded performance in the bias-guiding images.
Therefore, we demonstrate that our method represents the debiasing approach with wide applicability on real-world datasets.

To validate the proposed methods, the ablation study is provided in Section~\ref{sec:ablation_study} of the supplementary material.

\noindent{
\textbf{Evaluation on $\tilde{y}_\text{bias}$ assignment}}
We provide the quantitative evaluation on the proposed method of assigning the pseudo-bias labels on the bias-guiding and bias-contrary images via Eqs.~\ref{eq:bias score} and \ref{eq:pseudo bias label}, as shown in Table~\ref{tab: confusing score}.
The table includes the precision, recall, and F1 Score of the binary classification on the four datasets with varying ratios of bias severity.
Note that we consider identifying both the bias-guiding and bias-contrary images equally important in our framework, we first calculate each metric for both cases.
Afterwards, we add them and divide them by two in order to obtain the overall scores for classifying both bias-guiding and bias-contrary images.
As shown in Table~\ref{tab: confusing score}, the proposed method mentioned in Section~\ref{sec:3.1} achieves the reasonable performance on dividing the bias-guiding and bias-contrary images.
Therefore, providing these paired sets of images enables the effective generation of bias-swapped images in the swapping autoencoder described in Section~\ref{sec:bias-tailored swapping autoencoder}.

\begin{figure}[!t]
\centering
\begin{subfigure}[b]{\linewidth}
    \includegraphics[width=\linewidth]{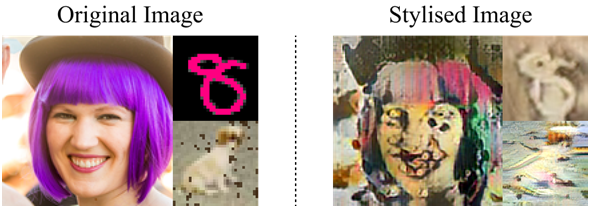}
    \vspace{-0.5cm}
    \caption{Generated images via Stylised ImageNet}
    \label{fig: stylised}
\end{subfigure}
\begin{subfigure}[b]{\linewidth}
    \includegraphics[width=\linewidth]{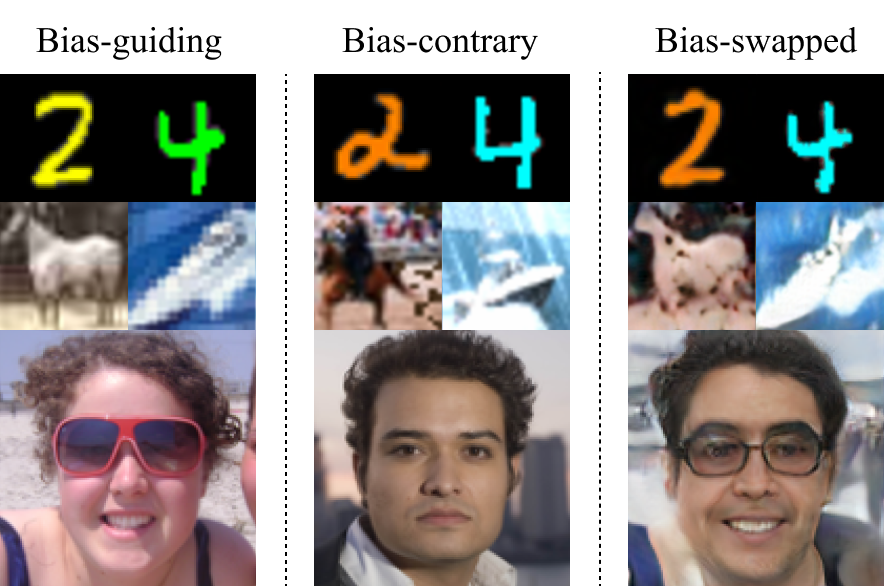}
    \vspace{-0.5cm}
    \caption{Generated images via BiaSwap}
    \label{fig: our example}
\end{subfigure}
\vspace{-0.5cm}
  \caption{Qualitative comparisons between the augmented images from (a) SIN and (b) our bias-tailored swapping augmentation. BiaSwap generates a more realistic and bias-aware image compared to SIN.}
\label{fig:cherry}
\vspace{-0.5cm}
\end{figure}

\subsection{Qualitative analysis}
\label{sec: qualitative}

\noindent\textbf{Generation of bias-swapped image}
Figure~\ref{fig:cherry}-(b) depicts a set of bias-guiding, bias-contrary, and the generated bias-swapped images for Colored MNIST, Corrupted CIFAR10, and bFFHQ in each row.
We observe that the bias-swapped images contain the bias attributes extracted from the bias-contrary images while maintaining the bias-irrelevant attributes from the bias-guiding images.
For example, our method translates a young female (first column) into a young male (third column) by reflecting the gender attribute from another young male (second column), while retaining the bias-irrelevant aspects, such as wearing sunglasses and smiling.

\begin{table}[t!]
\begin{center}
\scalebox{0.88}{
\begin{tabular}{c|c|c|c}
\toprule
 & Colored MNIST & BAR & bFFHQ\\
\midrule
\midrule
Vanilla &48.76& 49.59&74.86\\
\midrule
SIN & 40.79 & 50.51 & 69.86\\
\toprule
BiaSwap &\textbf{83.74}&\textbf{52.44}&\textbf{78.98}\\
\bottomrule
\end{tabular}
}
\end{center}
\vspace{-0.5cm}
\caption{Quantitative comparisons of unbiased test accuracy between BiaSwap and SIN. We utilize 99.0\% of bias-guiding images for training.}
\label{tab: style and ours}
\vspace{-0.5cm}
\end{table}

\noindent{\textbf{Comparison with stylised imagenet}}
Although existing augmentation-based methods have achieved the improved classification performance~\cite{yun2019cutmix, Walawalkar2020AttentiveCA, xie2019unsupervised}, they may suffer from generating unrealistic images. 
Some of the methods often leverage the simple image-level augmentation techniques to combine the two different images, resulting in unrealistic images compared to natural ones.
The recently proposed StylisedImageNet (SIN)~\cite{geirhos2018imagenettrained} utilizes the AdaIN-based style transfer to augment the ImageNet images with different textures in order to solve the texture bias.
However, as shown in Fig~\ref{fig:cherry}-(a), stylized results are much more unrealistic compared to the original images.
In contrast, our approach synthesizes realistic images in a more natural way, and we believe that realistic augmented images help debiasing more than unrealistic ones. 

To verify this, we compare the unbiased test accuracy of our method against that of SIN across the Colored MNIST, bFFHQ, and BAR datasets, in Table~\ref{tab: style and ours}.
We observe that SIN fails to learn the debiased representation on each dataset, and it may be caused by 1) the unrealistic augmented samples and 2) the augmentation without considering the bias attributes.
To be specific, in Figure~\ref{fig:cherry}-(a), stylized pink eight loses the original shape of eight, and the texture of a facial image is changed while the gender attribute remains unchanged.
On the other hand, BiaSwap only replaces the bias-relative attributes and generates visually plausible images, as shown in Figure~\ref{fig:cherry}-(b).

\begin{figure}[t!]
\begin{center}
\includegraphics[width=\linewidth]{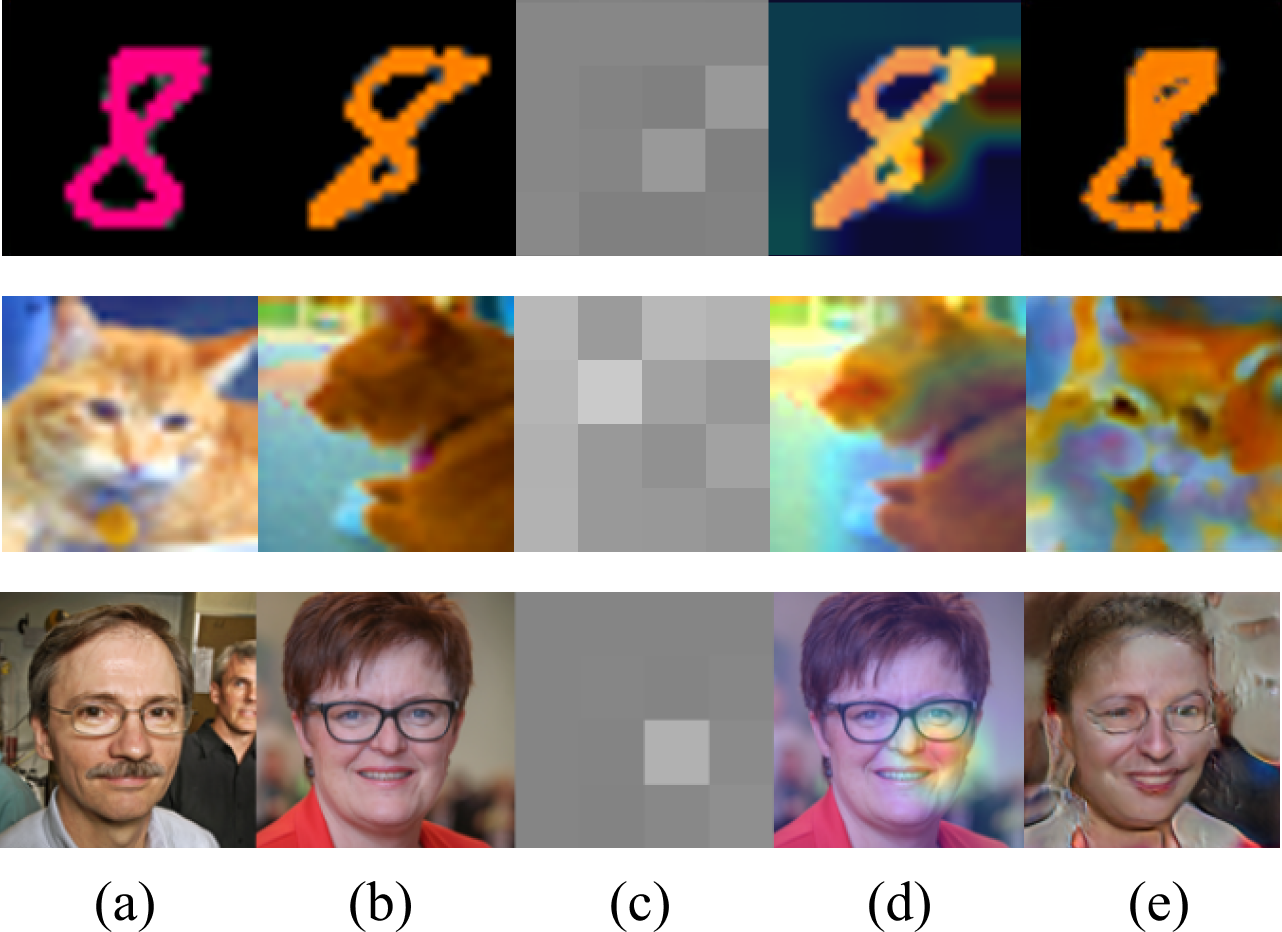}
\end{center}
\vspace{-0.5cm}
   \caption{Visualization of CAM utilized in our CAM-based patch sampler. 
   }
\label{fig: gradcam}
\vspace{-0.5cm}
\end{figure}

\noindent{\textbf{Visualization of CAM}}
In order from left to right in Figure~\ref{fig: gradcam}, bias-guiding sample, bias-contrary sample, CAM, CAM heatmap visualized on the image (b), and the bias-swapped image generated from BiaSwap are shown. 
Red regions in (d) correspond to the more discriminative region compared to the blue regions.
As intended, the highlighted regions of CAM mainly appear in the regions where the bias attributes are exploited, \textit{e.g.,} colors in Colored MNIST and face in bFFHQ.
Note that by exploiting the attributes of those attended regions in the biased classifier, our bias-tailored patch discriminator generates the images considering bias-relevant attributes.
For example, the cat in the (e) column contains the same corruption (\textit{i.e.,} saturate) as the one in column (b) while maintaining the overall shape of cat in column (a).

\section{Discussion and Conclusion}
In this paper, we propose a novel image translation-based debiasing approach which augments the realistic bias-contrary images for learning debiased representation.
Based on the assumption that bias attribute is easy-to-learn, we leverage the patch cooccurence discriminator integrated with CAM and the GCE loss to generate the image with its bias attributes translated from the bias-contrary images while maintaining the other bias-irrelevant visual aspects.
Extensive experiments demonstrate that our method successfully generates realistic bias-contrary images, achieving the state-of-the-art debiasing performance across diverse datasets.
We acknowledge that the perfect translation of bias in images remains challenging, particularly when the dataset contains a complex combination of bias attributes or the number of training images is limited.
However, we believe that our work can be viewed as a cornerstone of future debiasing works.

\vspace{0.2cm}
\small\noindent{\small\textbf{Acknowledgements}} This work was supported by Institute of Information \& communications Technology Planning \& Evaluation (IITP) grant funded by the Korea government(MSIT) (No. 2019-0-00075, Artificial Intelligence Graduate School Program(KAIST)).
This work was also supported by the National Research Foundation of Korea (NRF) grant funded by the Korean government (MSIT) (No. NRF-2019R1A2C4070420).
This work was also supported by Institute of Information \& communications Technology Planning \& Evaluation(IITP) grant funded by the Korea government(MSIT) (No. 2021-0-01778, Development of human image synthesis and discrimination technology below the perceptual threshold)

{\small
\bibliographystyle{unsrt}
\bibliography{main}
}

\clearpage


\appendix

\ificcvfinal\thispagestyle{empty}\fi
This material complements our paper with additional experimental results and their analysis. 
First of all, we present the ablation studies on the proposed modules of our framework, presented in Section~\ref{sec:ablation_study}.
Afterward, Section~\ref{sec:qual_confusing} provides the qualitative and quantitative analysis on the proposed pseudo-bias labels assignment.
In Section~\ref{sec:qual_samples}, we provide additional qualitative examples of augmented bias-swapped images generated by our proposed method, along with their class activation map (CAM)~\cite{zhou2015cam} visualization.
Section~\ref{sec:implementation} describes the implementation details, such as the settings for training and the construction of biased-FFHQ (bFFHQ) dataset.
Lastly, we provide a detailed explanation of the datasets and baselines we utilized in Section~\ref{sec:dataset}.

\section{Ablation study}
\label{sec:ablation_study}
This section demonstrates the effectiveness of our two main contributions, $1$) separation of bias-contrary and bias-guiding images and $2$) CAM-based patch sampling in the bias-tailored swapping autoencoder (SwapAE).
As explained in Sections~\ref{sec:3.1} and~\ref{sec:bias-tailored swapping autoencoder} of the main paper, the separation of contrary and guiding images encourages the swapping autoencoder to translate the bias-guiding image into the bias-swapped one by reflecting the bias-contrary attributes.
In addition, CAM obtained from the biased classifier enables the sampling of patches based on the highly discriminative (\textit{i.e.,} highly bias-related) regions, enforcing the bias-tailored patch discriminator to translate the visual styles from them.
To verify the effectiveness of such methods, we conduct the ablation studies on these two components denoted as \textbf{c1} and \textbf{c2} in Table~\ref{tab: ablation} and compare the accuracy of the unbiased test set over Colored MNIST and bFFHQ datasets.

As the separation is ablated, a pair of two guiding images become more frequently provided in the SwapAE for augmenting the new image, compared to the bias-guiding and bias-contrary pairs.
As a result, the translation between these guiding images generates another guiding image, which does not help to remove the dataset bias in the training distribution.
It is observed in Table~\ref{tab: ablation} that the model with \textbf{c1} ablated shows a critically degraded performance in an unbiased test set compared to the highest accuracies in the bias-guiding dataset on both Colored MNIST and bFFHQ.
In contrast, our model trained with \textbf{c1} achieves superior performance both in guiding and unbiased test set, demonstrating that the classifier benefits from the augmented images generated from (\textit{bias-guiding}, \textit{bias-contrary}) pairs.
When we ablate the second method \textbf{c2}, the patches are randomly sampled as exactly the same as the original patch co-occurrence discriminator proposed in the original paper~\cite{park2020swapping} does.
This simply transfers the overall style of a bias-contrary to a bias-guiding image without considering the regions of bias-attribute.
However, utilizing the CAM-based patch sampling enables the further optimized image translation by focusing on transferring the bias-related attributes in the image.
Table~\ref{tab: ablation} indicates the proposed method equipped with \textbf{c2} achieves the best accuracy on the unbiased test set of both datasets.

\begin{table}[!h]
\begin{center}
\setlength\tabcolsep{4.5pt}
\begin{tabular}{c|cc|cc}
\toprule
\multirow{2}{*}{Methods} & \multicolumn{2}{c|}{Bias-guiding} & \multicolumn{2}{c}{Unbiased} \\ 
\cmidrule{2-5}
&
Colored & \multirow{2}{*}{bFFHQ} & Colored & \multirow{2}{*}{bFFHQ} \
\\&MNIST&&MNIST&\\ 
\midrule \midrule
\multirow{1}{*}{BiaSwap w/o \textbf{c1}} & 99.92 & 99.2 & 43.04 & 45.0\\
\cmidrule{1-5}
BiaSwap w/o \textbf{c2} & 98.98 & 99.0 & 84.16 & 51.2\\
\cmidrule{1-5}
BiaSwap (Full) & 99.24 & 99.13 & \textbf{86.03} & \textbf{58.87}\\
\bottomrule
\end{tabular}
\vspace{-0.1cm}
\begin{tablenotes}
\item \textbf{c1} \ : Separation of bias-contrary and bias-guiding pairs.
\item \textbf{c2} \ : CAM-based patch sampling.
\end{tablenotes}
\end{center}
\vspace{-0.5cm}
\caption{Quantitative comparisons of our proposed method and its ablated versions on Colored MNIST and bFFHQ datasets. 
The separation of bias-contrary and bias-guiding pairs (\textbf{c1}), and CAM-based patch sampling (\textbf{c2}) are ablated.}
\label{tab: ablation}
\vspace{-0.5cm}
\end{table}

\section{Analysis on pseudo-bias label assignment}
\label{sec:qual_confusing}
As introduced in Section~\ref{sec:3.1} of the main paper, we utilize a bias score as well as the pseudo-bias label $y_\text{pseudo}$ to divide the entire training dataset into bias-guiding and bias-contrary samples.
To provide the qualitative and quantitative verification on the effectiveness of such division, this section consists of two parts, 1) qualitative examples classified as a bias-guiding and bias-contrary by the method and 2) quantitative evaluation of the robustness of $y_\text{pseudo}$ assignment on the diverse dataset setups, as supplementary to the Table~\ref{tab: confusing score} in the main paper.

\subsection{BAR images separated by pseudo-bias label}

\begin{figure*}
\begin{center}
\includegraphics[width=\linewidth]{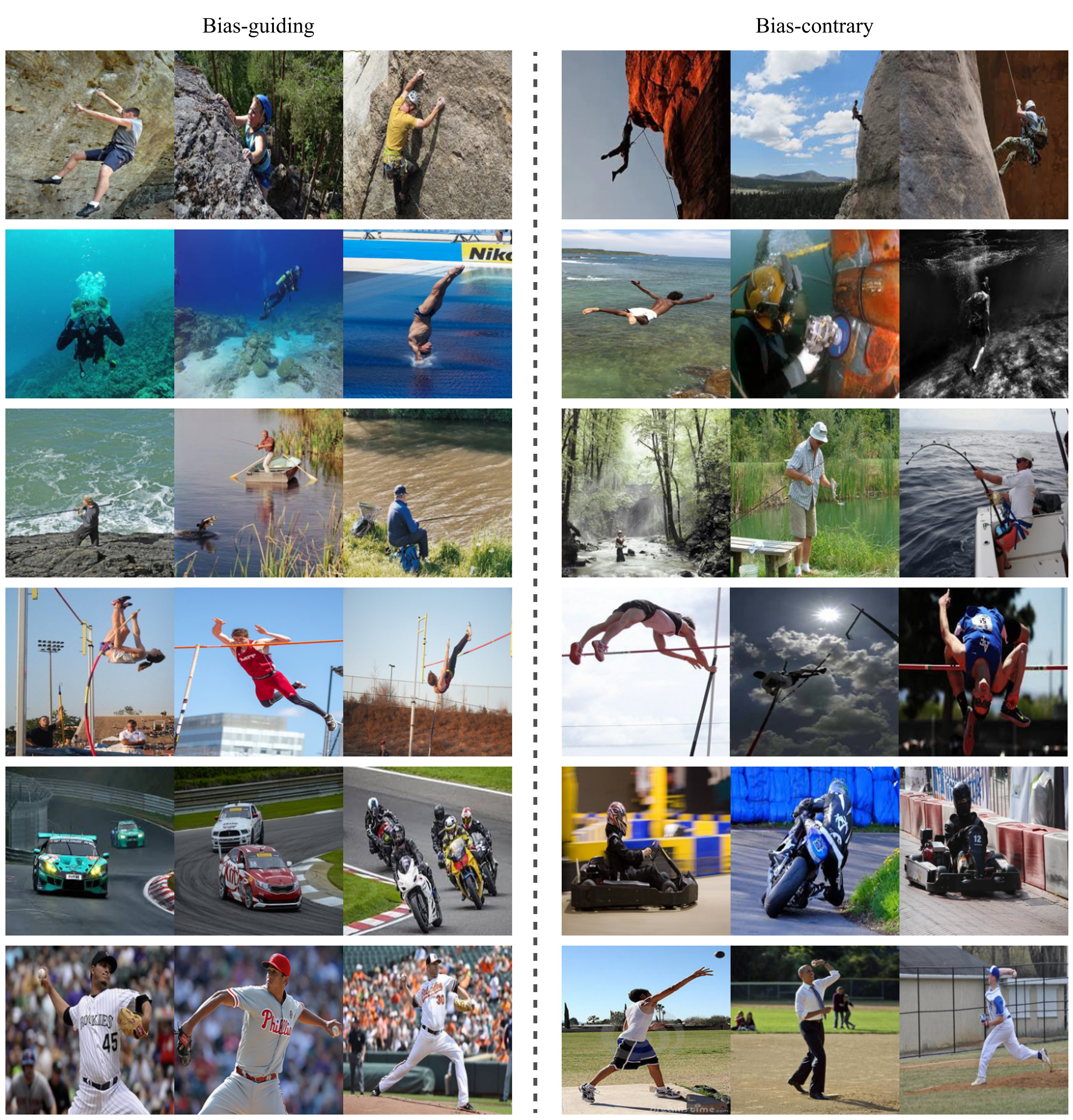}
\end{center}
\vspace{-0.5cm}
   \caption{Qualitative examples of divided images of BAR. Images on the left side represent the images classified as a bias-guiding sample based on our threshold. Images on the right side correspond to the images classified as a bias-contrary sample.}
\label{fig:threshold}
\vspace{0cm}
\end{figure*}

As mentioned in Section~\ref{sec:experimental setup} in the main paper, the training dataset of BAR only contains bias-guiding samples categorized by Nam \etal~\cite{nam2020learning}. 
However, even if we assume that all the samples have an unwanted correlation with the target label, there must exist a varying degree of bias between the training samples.
In other words, some of the samples can be more bias-guiding compared to the other images.
In this regard, the proposed dividing strategy can effectively capture this subtle difference and sorts the samples \textit{from easy to hard} one. 
To be specific, the images with the same ground-truth target labels can be sorted by their bias score, which represents how much the image includes the bias attributes.
Figure~\ref{fig:threshold} shows the exemplar images which are assigned to the bias-guiding (left columns) or bias-contrary (right columns) via our bias score and the threshold described in Eqs~\ref{eq:bias score} and~\ref{eq:pseudo bias label} in the main paper, respectively.
While unwanted correlations shown in the left columns are found in the most of training data, it turns out that some of the samples are relatively \textit{hard-to-learn}, where there exist less severe correlations between the bias attributes and the target label.
For instance, for the images labeled with climbing in the first row, most of the climbers are located on the rock which has an uneven texture and brown color. 
In this context, the examples with the sky in the most part of their backgrounds or with colors other than brown are classified as a bias-contrary sample. 
Similarly, for the images labeled with diving in the second row, divers are usually in the deep sea or taking a similar body motion.
However, examples on the right side are conflicting with the bias in that they contain a unique diving pose.
Some of them also are black-and-white pictures, which is uncommon in the training dataset.
For the fishing images on the right side of the third row, the fisher in the lake surrounded by the dense trees or the fisher near the river represents that such places are not the usual cases in the training distribution.
This implies that our method empirically well divides the samples based on the relative severity of the bias between the images.

\begin{table}[t!]
\begin{center}
\scalebox{0.88}{
\begin{tabular}{c|c|cccccc}
\toprule
Dataset & \% & Precision (\%) & Recall (\%) & F1 score (\%) \\
\midrule \midrule
\multirow{4}[4]{*}{\specialcell{Colored \\ MNIST}} 
& 95.0 & 97.09 & 98.43 & 93.55 \\ 
\cmidrule{2-5}
& 98.0 & 99.55 & 91.76 & 95.31 \\
\cmidrule{2-5}
& 99.0 & 97.54 & 92.12 & 94.74 \\
\cmidrule{2-5}
& 99.5 & 99.98 & 95.78 & 97.79 \\
\cmidrule{1-5}
\multirow{4}[4]{*}{\specialcell{Corrupted \\ CIFAR10}} 
& 95.0 & 68.34 & 80.66 & 72.56 \\ 
\cmidrule{2-5}
& 98.0 & 64.38 & 82.14 & 69.49 \\
\cmidrule{2-5}
& 99.0 & 60.7 & 87.28 & 66.13 \\
\cmidrule{2-5}
& 99.5 & 58.61 & 87.09 & 63.64 \\
\cmidrule{1-5}
bFFHQ & 99.0 & 65.52 & 70.62 & 67.70\\
\bottomrule
\end{tabular}
}
\vspace{-0.1cm}
\end{center}
\vspace{-0.5cm}
\caption{Quantitative evaluations on the $\tilde{y}_\text{bias}$ assignment via precision, recall, and F1 score metrics. We report the evaluation scores for 99.5\%, 99\%, 98\%, and 95\% of both Colored MNIST and Corrupted CIFAR10, and 99\% of bFFHQ, except BAR where no bias label is accessible.}
\label{tab: sup_confusing score}
\vspace{-0.5cm}
\end{table}

\subsection{Quantitative evaluation on pseudo-bias label}
Table~\ref{tab: sup_confusing score} demonstrates the quantitative evaluation scores of our proposed dividing method on the Colored MNIST, Corrupted CIFAR10, and bFFHQ datasets.
As our method works as a binary classifier to discriminate whether the images are bias-guiding or bias-contrary, we measure the precision, recall, and F1 score for both bias-guiding and bias-contrary classes on the unbiased test set.
Following the same evaluation protocol in Section~\ref{sec:Quantitative} of the main paper, we add the scores of bias-guiding and bias-contrary ones and divide them by two in order to obtain the overall scores.

Table~\ref{tab: sup_confusing score} indicates that the dividing method works well on classifying the bias-contrary as well as bias-guiding samples, achieving reasonable results in precision, recall, and F1 score.
In consequence, the robustness of the method enables to guarantee of the effective augmentation of bias-swapped images in the bias-tailored swapping autoencoder.

\section{Additional qualitative examples of bias-swapped images}
\label{sec:qual_samples}
\begin{figure}
\begin{center}
\includegraphics[width=\linewidth]{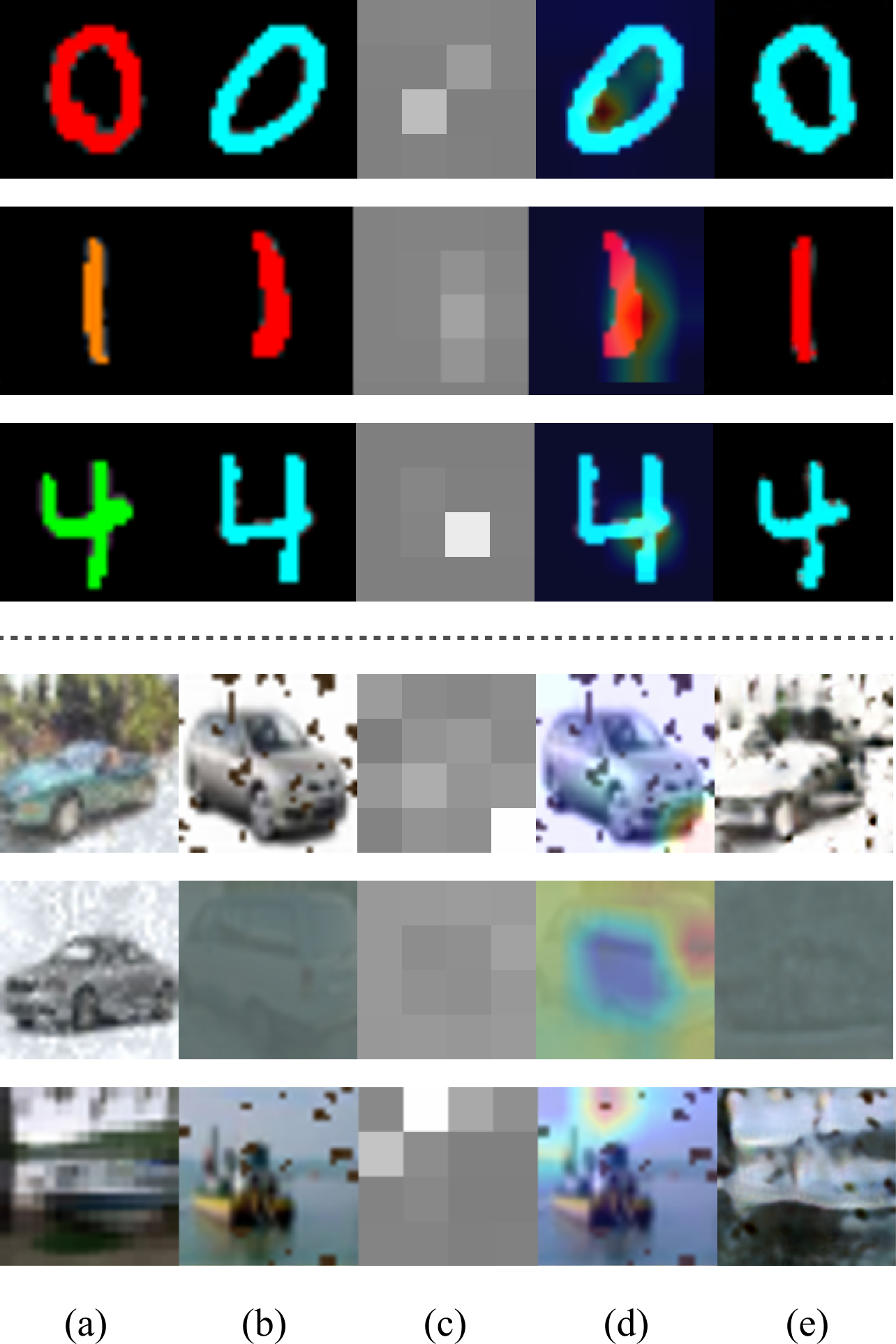}
\end{center}
\vspace{-0.5cm}
   \caption{Qualitative examples of generated images via BiaSwap on Colored MNIST (top), Corrupted CIFAR10 (middle) and bFFHQ (bottom).
   Each sample is listed in order of bias-guiding, bias-contrary, CAM on the bias-contrary, heatmap of CAM on the bias-contrary, and bias-swapped image.
   }
\label{fig:sup_cherry}
\vspace{0cm}
\end{figure}

To supplement Section~\ref{sec: qualitative} in the main paper, this section provides the additional qualitative results of the bias-swapped samples as well as their CAMs in Figure~\ref{fig:sup_cherry}.
In order from left to right, bias-guiding sample, bias-contrary sample, CAM, heatmap of CAM visualized on the bias-contrary image, and the bias-swapped image generated from BiaSwap are presented.
The first and second rows include the examples of Colored MNIST and Corrupted CIFAR10, respectively. 
Similar to the ones in the main paper, CAM heatmaps on the Colored MNIST samples show that the biased classifier mainly focuses on the regions where the bias-correlated colors are located.
For example, CAM described in the first row follows the region of blue colors in the digit.
For the samples of the Corrupted CIFAR10, our model properly transfers the bias attributes of the second column images into the ones in the first column, maintaining the bias-irrelevant visual aspects unchanged.
This results in the bias-swapped images shown in the last column.
For example, the corruption applied on the car in the second column is transferred into another car in the first column, while the shape of the first column car is maintained in the generated bias-swapped car in the last column.

\section{Implementation details}
\label{sec:implementation}
This section provides the specific values of the threshold for separation of bias-guiding and bias-contrary groups over each dataset.
Afterward, we provide the detailed architecture of two main networks, bias-tailored swapping autoencoder and debiased classifier.
In addition, we provide the training details, such as hyper-parameters for each objective function, over the dataset we utilized.

\noindent \textbf{Threshold for division}
To separate the training samples into bias-contrary and bias-guiding sets in an unsupervised manner, we utilize the mean value of confusing scores of the images as our threshold in each dataset, as described in Eq.~\ref{eq:pseudo bias label} in the main paper.
Such threshold values correspond to $0.0358$, $0.0903$, $0.0232$, and $0.008$ for Colored MNIST, Corrupted CIFAR10, BAR, and bFFHQ, respectively.

\noindent \textbf{Bias-tailored swapping autoencoder}
We follow the original network architecture of the encoder, decoder, and discriminator proposed in Park \etal~\cite{park2020swapping} to maintain its image translation performance.
However, to design a CAM-based patch sampling for the co-occurrence discriminator as proposed in Section~\ref{sec:bias-tailored swapping autoencoder} in the main paper, we sample the patches using the patch-wise probability based on the CAM of the biased classifier, instead of random sampling.
For the biased classifier, we use a multi-layer perceptron (MLP) with three hidden layers for Colored MNIST and ResNet-18~\cite{He2015resnet} for the rest of the datasets.
The classifier is trained with the GCE loss with hyperparameter $q$ of $0.7$.
As the parameters of the classifier are \textit{not} jointly optimized with those of bias-tailored swapping autoencoder, the classifier is fully trained to be biased.
We train and evaluate the biased classifier with the size of $28\times28$ and $32\times32$ images for Colored MNIST and Corrupted CIFAR10, and $128\times128$ for BAR and bFFHQ datasets.
Each channel of the images are normalized with the mean of $(0.5, 0.5, 0.5)$ and the standard deviation of $(0.5, 0.5, 0.5)$. 
All other details are identical to the training of the debiased classifier. 
To train the autoencoder, we set the hyper-parameters for each loss functions as $\lambda_{\text{recon}}=\lambda_{\text{GAN,recon}}=\lambda_{\text{GAN,swap}}=\lambda_{\text{CooccurGAN}}=1$.
To prevent the patch discriminator from only sampling the same single patch due to the high probability close to one, we utilize the temperature scaling with $\tau=10$, for smoothing the probability.
Both the swapping autoencoder and the classifier are trained by using an  Adam~\cite{kingma2014adam} optimizer with $\beta_1=0$ and $\beta_2=0.99$.

\noindent \textbf{Debiased classifier}
After we fully optimize the bias-tailored swapping autoencoder, we augment the dataset using the pairs of bias-guiding and contrary images given from the threshold.
Given these additional images, which we call \textit{bias-swapped} images in the main paper, we train an MLP with three hidden layers for Color MNIST and ResNet-18 for the rest of the datasets.
We train and evaluate the classifier with the size of $28\times28$, $32\times32$, $128\times128$, and $224\times224$ images for Colored MNIST, Corrupted CIFAR10, bFFHQ, and BAR datasets, respectively.
We use Adam optimizer with $\beta_1=0.9$ and $\beta_2=0.999$, and learning rate as $0.001$.
We use a batch size of 256 and train a classifier for 200 epochs for Color MNIST, Corrupted CIFAR10, BAR, and bFFHQ datasets.

\section{Datasets and Baselines}
\label{sec:dataset}
\subsection{Datasets}
\noindent \textbf{Corrupted CIFAR10}
As proposed in Hendrycks and Dietterich~\cite{hendrycks2018benchmarking}, we apply the certain type of texture corruptions onto the CIFAR10 dataset~\cite{Krizhevsky09}.
Among 15 types of corruptions, we utilize the \textit{Snow, Frost, Fog, Brightness, Contrast, Spatter, Elastic, JPEG, Pixelate}, and \textit{Saturate} in our paper.
Such corruptions are applied with the strong correlation with the original classes of CIFAR10 dataset, which are \textit{Plane, Car, Bird, Cat, Deer, Dog, Frog, Horse, Ship, and Truck.}
In addition, we utilize the corruptions with the highest degree of severity (\textit{i.e.}, 4) in our dataset.

\noindent \textbf{bFFHQ}
We newly construct the biased FFHQ dataset (bFFHQ) which has a strong correlation between the age (target) and gender (bias), based on the Flickr-Faces-HQ (FFHQ) dataset~\cite{karras2019style}. 
FFHQ consists of $70,000$ images at $1024\times1024$ resolution and contains the considerable variation of human faces in terms of age, ethnicity, and image background. 
Each face contains different attributes including head pose, gender, age, mustache, glasses, and emotion. 
Among these attributes, we utilize the \textit{age} and \textit{gender} attributes. 
To be specific, the attribute `young' (\textit{i.e.}, aged $10-29$) is highly correlated with `women' and `old' (\textit{i.e.}, aged $40-59$) is connected with `men'. 
Among the total $70,000$ data, $19,200$ samples are utilized as the training dataset according to the criteria of unwanted correction, and 2,000 unbiased samples that each attribute is uniformly distributed are utilized as an evaluation set.

\noindent \textbf{BAR}
This dataset includes the images which have a correlation between human actions and backgrounds, which is curated by Nam \etal~\cite{nam2020learning}.
It does not have the ground-truth bias label, unlike other datasets.

\subsection{Baselines}
\noindent \textbf{ReBias}
ReBias~\cite{bahng2019rebias} assumes the \textit{texture} as the unwanted bias type and exploits the biased classifier with a limited kernel size in order to mainly capture the texture attributes from the images.
By learning the representations statistically independent from such texture representations, ReBias achieves robust classification accuracies against the texture bias.
We train ReBias with the same training protocol as suggested in the original paper including network architecture and the hyper-parameters.
Note that for Colored MNIST, ReBias utilizes the convolutional network for capturing the texture cues, while other baselines including ours exploit the MLP with three hidden layers.

\noindent \textbf{LfF}
As mentioned in Sections~\ref{sec:introduction} and~\ref{sec: related work} of the main paper, LfF assumes the general characteristic of bias as ``easy-to-learn" and proposes the re-weighting-based debiasing method based on the GCE loss.
To the best of our knowledge, this work first learns the debiased representation without any prior assumption on the bias type.
We follow the official implementation setups of LfF, except for the network architecture of ResNet-20 for the Corrupted CIFAR10 dataset.
As a fair comparison, we utilize the ResNet-18 architecture for all the baselines including LfF.

\noindent \textbf{SIN}
As mentioned in Section~\ref{sec: qualitative} of the main paper, we utilize SIN as another baseline for validating the importance of realistic image generation in dataset augmentation.
For the augmentation of datasets we utilize, we follow the official implementation of SIN and only replace the original ImageNet dataset with each biased dataset. Style images are identical to official ones.

\end{document}